\documentclass{article}
\usepackage[utf8]{inputenc}
\usepackage{arxiv}
\usepackage{amsmath,amssymb,mathtools}
\usepackage{graphicx}
\usepackage{booktabs}
\usepackage{hyperref}
\usepackage{cleveref}
\usepackage{caption}
\usepackage{xcolor}

\renewcommand{\headeright}{}
\renewcommand{\undertitle}{}

\title{ConstraintBench: Benchmarking LLM Constraint Reasoning on Direct Optimization}

\author{
  Joseph Tso, Preston Schmittou, Quan Huynh, Jibran Hutchins \\
  Haladir Research Team \\
  \texttt{\{joseph,preston,quan,jibran\}@haladir.com}
}

\date{}

\begin{document}

\maketitle

\begin{abstract}
Large language models are increasingly applied to operational decision-making where the underlying structure is constrained optimization. Existing benchmarks evaluate whether LLMs can formulate optimization problems as solver code, but leave open a complementary question. Can LLMs directly produce correct solutions to fully specified constrained optimization problems without access to a solver? We introduce ConstraintBench, a benchmark for evaluating LLMs on direct constrained optimization across 10 operations research domains, with all ground-truth solutions verified by the Gurobi solver. Each task presents a natural-language scenario with entities, constraints, and an optimization objective; the model must return a structured solution that a deterministic verifier checks against every constraint and the solver-proven optimum. We evaluate six frontier models on 200 tasks and find that feasibility, not optimality, is the primary bottleneck. The best model achieves only 65.0\% feasibility, yet feasible solutions average 89 to 96\% of the Gurobi-optimal objective. No model exceeds 30.5\% on joint feasibility and optimality within 0.1\% of the solver reference. Per-domain analysis shows large variation in difficulty, with average feasibility spanning from 85.0\% in the facility location domain to 0.8\% in the crew assignment domain. Further, systematic failure modes include duration constraint misunderstanding, entity hallucination, and a feasibility-optimality decoupling in facility location and vehicle routing where models achieve high feasibility but 0\% optimality. ConstraintBench and all evaluation infrastructure will be publicly released.
\end{abstract}

\section{Introduction}
\label{sec:introduction}

Constrained optimization underlies a broad range of business operations. Logistics companies solve facility location problems, hospitals solve nurse scheduling problems, and portfolio managers solve quadratic programs under risk limits. Large language models are increasingly deployed in settings where these decision problems arise, from planning and scheduling to allocation and pricing~\cite{anthropic2025vend1, anthropic2025vend2}, yet rigorously measuring whether LLMs make correct constrained decisions has remained difficult. The combinatorial search space is too large for humans to verify by inspection, and without a formal solver providing ground truth, there is no reference against which to compare.

A growing body of work benchmarks LLMs on optimization formulation, translating natural-language problem descriptions into executable solver code that is then run to check whether the resulting objective matches a reference value (NL4Opt~\cite{ramamonjison2022nl4opt}, IndustryOR~\cite{huang2024orlm}, Mamo~\cite{huang2024mamo}, OptMATH~\cite{lu2025optmath}, ComplexOR~\cite{xiao2024complexor}, OptiBench~\cite{yang2025optibench}). These benchmarks answer an important question, whether a model can write correct solver code, but leave a complementary question unaddressed: Can a model directly produce correct decisions for a fully specified optimization problem, without delegating to a solver at inference time? This question tests fundamentally different capabilities, such as constraint reasoning and combinatorial search rather than mathematical modeling and code generation.

This paper focuses on this direct-solution setting. Our contributions are as follows.

\begin{enumerate}
    \item We introduce ConstraintBench, a benchmark for evaluating LLMs on direct constrained optimization across 10 operations research (OR) domains with 200 verified problem-solution pairs. The problems focus on evaluating whether models can reason through constraint interactions and produce feasible, near-optimal solutions as structured outputs.
    \item We provide detailed per-domain analysis of the performance of six frontier models, revealing that producing feasible solutions is the primary bottleneck for most models, with domain-level feasibility ranging from 0.8\% to 85.0\%.

    \item We characterize the common failure modes for these tasks. These include duration constraint misunderstanding, entity hallucination, and a feasibility-optimality decoupling where models achieve high constraint satisfaction but 0\% optimality.

    \item We eliminate the data-quality problem that affects existing benchmarks. Rather than manual annotation, ground-truth solutions are computed using Gurobi, thus rendering every task as verifiably feasible and optimal prior to inclusion. 

\end{enumerate}

\section{Related Work}
\label{sec:related}

\subsection{LLM Benchmarks for Optimization Modeling}
\label{sec:related:benchmarks}

Several benchmarks evaluate LLMs on formulation, translating natural-language optimization problems into executable solver code. NL4Opt~\cite{ramamonjison2022nl4opt} introduced 1,101 annotated linear programming word problems, with 289 test instances commonly used for evaluation. IndustryOR~\cite{huang2024orlm} provides 100 real-world OR problems spanning linear, integer, mixed-integer, nonlinear, and other problem types. Mamo~\cite{huang2024mamo} offers a broader set of mathematical modeling problems with solver verification. OptMATH~\cite{lu2025optmath} proposes a scalable bidirectional data synthesis framework producing a larger-scale formulation benchmark. ComplexOR~\cite{xiao2024complexor} targets complex multi-step OR problems requiring chain-of-experts reasoning. OptiBench~\cite{yang2025optibench} measures formulation accuracy with Socratic-style refinement.

All of these benchmarks share a common evaluation pattern in which the LLM produces solver code, the code is executed, and the resulting objective is compared to a ground-truth value. ConstraintBench differs on three axes. First, it evaluates direct solution rather than formulation, meaning the model must produce decisions, not code. Second, evaluation is at the constraint level (every constraint checked individually, with violation diagnostics) rather than binary objective-match. Third, ground truth is solver-verified rather than human-annotated, addressing the data-quality problems reported in the literature (e.g., OptiMind~\cite{zhang2025optimind}) for IndustryOR and OptMATH. A detailed comparison is provided in \Cref{tab:relatedwork}.

\subsection{LLM Systems for Optimization}
\label{sec:related:systems}

Beyond benchmarks, several systems improve LLM performance on formulation tasks. OptiMUS~\cite{ahmaditehnizi2024optimus, ahmaditehnizi2024optimus03} uses agentic frameworks to iteratively formulate, debug, and solve optimization problems at scale. OptiMind~\cite{zhang2025optimind} teaches LLMs to reason like optimization experts through class-specific error analysis and solver feedback. SIRL~\cite{chen2025sirl} uses solver-informed reinforcement learning to ground LLM formulations in solver feedback. These are systems that improve formulation accuracy; ConstraintBench measures a different capability (direct constraint reasoning) and is complementary.

\subsection{LLMs for Combinatorial Optimization}
\label{sec:related:combinatorial}

A separate line of work uses LLMs as heuristic generators or direct solvers for combinatorial problems. FunSearch~\cite{romeraparedes2024funsearch} pairs an LLM with an evolutionary pipeline to discover novel heuristics for bin packing and the cap set problem, demonstrating that LLMs can contribute to mathematical discovery. ReEvo~\cite{ye2024reevo} introduces Language Hyper-Heuristics that combine evolutionary search with LLM-generated reflective feedback, achieving state-of-the-art results across TSP, CVRP, and bin packing while being more sample-efficient than FunSearch. Work on directly prompting LLMs for TSP~\cite{masoud2024tsp} evaluates zero-shot, few-shot, and chain-of-thought approaches, while Self-Guiding Exploration~\cite{iklassov2024selfguiding} uses an exploration-decomposition-refinement pipeline for multi-step combinatorial solving.

ConstraintBench is related but distinct. Rather than evaluating whether LLMs can discover novel heuristics or solve well-studied combinatorial problems in isolation, it measures whether LLMs can navigate realistic multi-constraint optimization scenarios where constraint interactions, not just objective optimization, are the primary challenge.

\begin{table}[ht]
\centering
\small
\caption{Related work comparison.}
\label{tab:relatedwork}
\begin{tabular}{llllll}
\toprule
Benchmark & Task & Domains & \# Problems & Verification & Data Quality \\
\midrule
NL4Opt~\cite{ramamonjison2022nl4opt} & Formulation & LP & 1,101 (289 test) & Objective match & Human-annotated \\
IndustryOR~\cite{huang2024orlm} & Formulation & Mixed & 100 & Objective match & Human-annotated \\
Mamo~\cite{huang2024mamo} & Formulation & Mixed & -- & Objective match & Human-annotated \\
OptMATH~\cite{lu2025optmath} & Formulation & Mixed & Large-scale & Objective match & Synthetic + human \\
ComplexOR~\cite{xiao2024complexor} & Formulation & MILP & -- & Objective match & Human-annotated \\
OptiBench~\cite{yang2025optibench} & Formulation & Mixed & -- & Objective match & Human-annotated \\
\textbf{ConstraintBench} & \textbf{Direct solution} & \textbf{10 OR domains} & \textbf{200} & \textbf{Constraint-level + gap} & \textbf{Solver-verified} \\
\bottomrule
\end{tabular}
\end{table}

\section{Preliminaries}
\label{sec:preliminaries}

\subsection{Mixed-Integer Programming}
\label{sec:prelim:mip}

We model each task as a mixed-integer program with linear constraints and a linear objective. A solution $x$ is considered feasible if it satisfies all constraints and integrality requirements. Given the solver-proven optimum $x^*$ with objective value $f(x^*)$, the optimality gap of a proposed solution $x$ is defined as
\begin{equation}
\text{gap}(x) = \frac{f(x) - f(x^*)}{|f(x^*)|}
\label{eq:gap}
\end{equation}
For minimization problems, this definition applies directly; for maximization, we invert the sign. In continuous domains, floating-point arithmetic can produce small negative gaps where a model's solution appears to slightly outperform the solver reference; we treat $|\text{gap}(x)| < \epsilon$ as optimal for tolerance $\epsilon$. We define two optimality metrics. For a feasible solution $x$, the \emph{objective quality} is $(1 - |\text{gap}(x)|) \times 100$, representing the percentage of the solver-optimal objective value achieved. The \emph{near-optimality rate} is $P(|\text{gap}(x)| < \epsilon \mid x \text{ is feasible})$, the fraction of feasible solutions within tolerance $\epsilon$ of the optimum.

\subsection{Solvers and Verification}
\label{sec:prelim:solvers}

We use the Gurobi Optimizer~\cite{gurobi2024} (Gurobi) throughout this work. Gurobi solves MIPs via branch-and-bound with cutting planes, LP relaxation, and presolve techniques. Critically for this benchmark, when Gurobi reports an optimal solution, it has mathematically proven that no feasible solution with a better objective value exists. Further, when Gurobi reports infeasibility, it can extract an Irreducible Infeasible Subsystem (IIS), a minimal subset of constraints that cannot be simultaneously satisfied. Leveraging these abilities provides formal grounding and demonstrates that solutions are solver-proven, rather than heuristically estimated.

\subsection{Example Domain: Facility Location}
\label{sec:prelim:facility}

To give a concrete sense of what models must reason about, we present the MIP formulation for one domain. In the facility location problem, we decide which facilities to open and how to assign customers to open facilities.

\textbf{Decision variables.} Binary variables $y_j \in \{0,1\}$ indicate whether facility $j$ is opened. Continuous variables $x_{ij} \in [0,1]$ represent the fraction of customer $i$'s demand served by facility $j$.

\textbf{Objective.} Minimize total cost, $\sum_j f_j y_j + \sum_{i,j} v_j d_i x_{ij}$, where $f_j$ is the fixed cost of opening facility $j$, $v_j$ is the per-unit variable cost, and $d_i$ is customer $i$'s demand.

\textbf{Constraints.} Each customer's demand must be fully satisfied, $\sum_j x_{ij} = 1 \; \forall i$. Assignments require open facilities, $x_{ij} \leq y_j \; \forall i,j$. Facility capacity must not be exceeded, $\sum_i d_i x_{ij} \leq C_j y_j \; \forall j$. Customers may only be assigned to facilities within a maximum distance ($x_{ij} = 0$ if $\text{dist}(i,j) > D_{\max}$). Additional logical constraints (e.g., ``if facility A is opened, facility B must also be opened,'' or ``facilities C and D cannot both operate'') add combinatorial structure.

The model must reason about trade-offs between fixed costs, variable costs, capacity utilization, and distance constraints, while simultaneously satisfying all logical constraints. This illustrates the challenge ConstraintBench poses. The mathematical structure is not especially complex, but the interaction of multiple constraint types in a realistic scenario requires careful combinatorial reasoning.

\section{ConstraintBench}
\label{sec:constraintbench}

\subsection{Domain Coverage}
\label{sec:cb:domains}

ConstraintBench spans 10 operations research domains selected to cover the major axes of variation in constrained optimization: variable type (binary, integer, continuous, and mixed), constraint structure (capacity, precedence, temporal, logical), and objective class (cost minimization, profit maximization, makespan minimization). Table~\ref{tab:domains} summarizes each domain.

The domains range from tightly constrained combinatorial problems to continuous allocation tasks. Crew assignment requires continuous hour allocations subject to coverage requirements, workload balance limits, and qualification constraints that interact densely, creating a structure where satisfying any single constraint is easy but satisfying all simultaneously is difficult. Production mix involves integer production quantities under capacity and resource constraints with relatively independent constraint interactions. Domains like vehicle routing and job-shop scheduling combine binary or mixed decision variables with temporal constraints (time windows, operation precedence) that create long dependency chains, while facility location pairs binary open/close decisions with continuous assignment variables linked by capacity and distance constraints alongside ad hoc logical requirements (e.g., ``if facility A opens, facility B must also open'').

We note that ensuring comparable mathematical difficulty across domains is not feasible in general, and the 10 domains are not calibrated to a uniform difficulty level. Rather, the design goal is breadth of constraint structure. Domains with independent constraints test whether models can satisfy many requirements in parallel; domains with interacting constraints test whether models can reason about how one decision propagates through the feasible region. As shown in Section 5.3, this produces substantial variation in empirical difficulty, with average feasibility spanning two orders of magnitude across the 10 domains.

Each domain is implemented as a Pydantic schema defining the solution space, a Gurobi model builder that translates each scenario into a formal MIP, and an independent verifier that checks every constraint and recalculates the objective from raw decision variables. The model under evaluation receives only the natural-language prompt and the required output schema; it does not receive the Gurobi model, the reference solution, or any feedback.

\begin{table}[ht]
\centering
\small
\caption{ConstraintBench domains.}
\label{tab:domains}
\begin{tabular}{llll}
\toprule
Domain & Decision Variables & Objective & Variable Type \\
\midrule
Order Fulfillment & Source allocation per order-product pair & Minimize total cost & Integer \\
Production Mix & Units to produce per product & Maximize total profit & Integer \\
Shift Scheduling & Employee-shift assignments & Minimize total cost & Binary \\
Crew Assignment & Worker-task hour allocations & Minimize total cost & Continuous \\
Job-Shop Scheduling & Operation start times and machine assignments & Minimize makespan & Mixed \\
Project Planning & Task start/finish periods & Minimize project duration & Integer \\
Vehicle Routing & Route arc selections per vehicle & Minimize total distance & Binary \\
Bin Packing & Item-bin assignments & Minimize bins used / cost & Binary \\
Portfolio Optimization & Capital allocation per asset & Maximize expected return & Continuous \\
Facility Location & Facility open decisions and customer assignments & Minimize total cost & Mixed \\
\bottomrule
\end{tabular}
\end{table}

\subsection{Task Generation}
\label{sec:cb:generation}
Tasks are generated via a three-stage pipeline consisting of seed generation, LLM-driven scenario creation, and Gurobi solving.

\textbf{Seed generation.} Each task begins with a seed drawn from a five-dimensional combinatorial product of industry contexts (${\sim}45$ per domain), company scales (5), urgency levels (5), geographic regions (5), and domain specializations (5). This yields approximately 28{,}000 unique seeds per domain. Seeds are generated deterministically via a seeded random number generator (\texttt{seed=42}) using mixed-radix decomposition over the product space, with sampling without replacement. A composed seed takes the form ``E-commerce electronics retailer fulfilling online orders, operating as a mid-size regional company, during seasonal peak demand, in US domestic operations, with just-in-time inventory requirements.''

\textbf{LLM-driven scenario creation.} A LangGraph agent with access to four tools (sandboxed Python execution, structural validation, a task-saving scratchpad, and Gurobi feasibility checking) generates a concrete optimization scenario from each seed. The agent may iterate up to 9 times, with Gurobi's IIS feedback on infeasible attempts appended to the prompt for subsequent iterations.

\textbf{Gurobi solving.} Each generated task is solved by Gurobi to confirm feasibility and extract the optimal solution with a proof of optimality. Tasks where Gurobi cannot prove optimality within the time limit are discarded. The benchmark samples 20 tasks per domain (200 total) from the feasible, solver-verified instances.

\subsection{Evaluation Protocol}
\label{sec:cb:eval}

All models are evaluated under standardized conditions. Each model receives the natural-language problem prompt and must return a structured JSON response conforming to the domain-specific schema in a single turn. The model does not receive the reference solution, the Gurobi model, or any feedback.
Evaluation hyperparameters are listed in Appendix~\ref{app:eval_hyperparams}.

Verification of model generation is as follows:

\textbf{Step 1, response parsing and structural validation.} The model's structured output is parsed into domain-specific decision variables. Parse failures are recorded as errors and counted as infeasible. The verifier then checks for unknown entity IDs, negative quantities, missing required fields, and other schema violations.

\textbf{Step 2, constraint checking.} Every constraint is evaluated against the proposed solution, producing a binary satisfied/violated result for each constraint. This includes both structural constraints inherent to the domain (e.g., each customer assigned to exactly one facility) and the custom constraints specified in the prompt (e.g., ``facility f1 must remain open''). Each violated constraint produces a human-readable diagnostic string.

\textbf{Step 3, independent metric calculation.} The verifier re-derives the objective value from the model's raw decision variables using the same mathematical formulation as the Gurobi model. Critically, the model's self-reported objective value is never trusted.

\textbf{Step 4, optimality gap computation.} The independently calculated objective is compared against the Gurobi-proven reference. A solution is classified as feasible if all constraints are satisfied (zero violations), and optimal if feasible and within 0.1\% of the Gurobi reference objective. There is no partial credit. A solution violating a single constraint receives the same feasibility classification as one violating all constraints.

Throughout, errors (API failures, parse failures) are counted as infeasible with denominator equal to all 200 tasks, ensuring that models are not rewarded for failing to produce parseable responses.

\section{Experiments}
\label{sec:experiments}

\subsection{Experimental Setup}
\label{sec:exp:setup}

We evaluate six frontier models from three providers. GPT-5.2-pro and GPT-5.2 (OpenAI), Claude Opus 4.6 and Claude Opus 4.5 (Anthropic), o4-mini (OpenAI), and Gemini 3 Pro Preview (Google). All models are evaluated on the full 200-task benchmark under the conditions described in \Cref{sec:cb:eval}.

\subsection{Aggregate Results}
\label{sec:exp:aggregate}

\Cref{tab:aggregate} presents aggregate results across all 200 tasks.

\begin{table}[t]
\centering
\caption{Aggregate results (feasibility, objective quality, and \% optimal).}
\label{tab:aggregate}
\vspace{6pt}
\begin{tabular}{lrrrr}
\toprule
Model & Feasibility (\%) & Obj.\ Quality (\%) & \% Optimal & Errors \\
\midrule
GPT-5.2-pro & 65.0 & 95.2 & 30.5 & 1 \\
GPT-5.2 & 64.0 & 93.8 & 21.0 & 1 \\
Claude Opus 4.6 & 49.0 & 96.0 & 21.5 & 13 \\
Claude Opus 4.5 & 50.5 & 94.0 & 12.5 & 5 \\
o4-mini & 50.5 & 89.0 & 8.5 & 2 \\
Gemini 3 Pro Preview & 44.5 & 94.8 & 8.5 & 0 \\
\bottomrule
\end{tabular}
\end{table}

The results demonstrate that feasibility, not optimality, is the primary bottleneck. The best-performing model (GPT-5.2-pro) achieves only 65.0\% feasibility, meaning over a third of its solutions violate at least one constraint. Yet among feasible solutions, objective quality is consistently high (95.2\% for GPT-5.2-pro, 96.0\% for Claude Opus 4.6), indicating that when models do find the feasible region, they tend to find good solutions within it.

This asymmetry reveals a significant capability gap. Models struggle with constraint satisfaction rather than objective optimization. The gap between the high objective quality (${\sim}$89 to 96\%) and the low joint near-optimality rates (${\sim}$8.5 to 30.5\%) underscores that finding the feasible region is the harder challenge.

However, the high optimality percentages among feasible solutions should be interpreted with care. This pattern is consistent with a well-known property of combinatorial optimization. For many NP-hard problems, polynomial-time algorithms can find solutions within a constant factor of the optimum, but finding the exact optimum remains intractable~\cite{williamson2011approximation}. Achieving 90 to 95\% of the optimal objective is often straightforward once a feasible solution is found, while closing the remaining gap requires navigating subtle trade-offs. Our strict 0.1\% threshold for ``\% Optimal'' reflects this. Even at ${\sim}$95\% average objective quality, only 30.5\% of all tasks achieve solutions within 0.1\% of the Gurobi reference (see \Cref{sec:exp:decoupling} for per-domain analysis of where this gap is concentrated).

It is also notable that model scale does not straightforwardly predict performance. The smaller o4-mini matches Claude Opus 4.5 on feasibility (50.5\%) but achieves lower objective quality among feasible solutions (89.0\% vs.\ 94.0\%), while Claude Opus 4.6 achieves higher objective quality (96.0\%) than GPT-5.2 (93.8\%) despite lower feasibility (49.0\% vs.\ 64.0\%). This suggests that constraint reasoning may be a capability axis partially independent of general model capability.

\subsection{Per-Domain Analysis}
\label{sec:exp:perdomain}

\Cref{tab:perdomain} summarizes per-domain feasibility and joint optimality; domains are ordered from easiest to hardest by feasibility.

The difficulty spectrum spans two orders of magnitude, from 85.0\% average feasibility in facility location and 83.3\% in production mix down to 8.3\% in project planning and 0.8\% in crew assignment. This dynamic range is evidence that the benchmark captures meaningful variation in constraint reasoning difficulty.

Model rankings are unstable across domains. o4-mini achieves the highest feasibility on order fulfillment (90.0\%) but the lowest on shift scheduling (55.0\%); Claude Opus 4.6 achieves 85\% to 90\% on production mix and facility location but collapses to 10\% on bin packing and vehicle routing. No single model dominates across all domains, suggesting that different models have developed different constraint reasoning strengths.

Three domains (facility location, vehicle routing, and crew assignment) achieve 0\% joint optimality across all six models, despite facility location achieving 85.0\% average feasibility (and one model at 100\%). We analyze this feasibility-optimality decoupling in \Cref{sec:exp:decoupling}.

\begin{table*}[t]
\centering
\small
\caption{Per-domain feasibility and joint optimality (\%).}
\label{tab:perdomain}
\resizebox{\textwidth}{!}{%
\begin{tabular}{lrrrrrrrrrr}
\toprule
Model & Fac.\ Loc.\ & Prod.\ Mix & Order Ful.\ & Job Shop & Shift Sch.\ & Portfolio & Vehicle & Bin Pack & Project & Crew \\
\midrule
\multicolumn{11}{l}{\textbf{Feasibility (\%)}} \\
\midrule
GPT-5.2-pro & 100.0 & 95.0 & 70.0 & 95.0 & 85.0 & 65.0 & 65.0 & 70.0 & 5.0 & 0.0 \\
GPT-5.2 & 85.0 & 90.0 & 80.0 & 80.0 & 80.0 & 65.0 & 80.0 & 50.0 & 25.0 & 5.0 \\
Claude Opus 4.6 & 90.0 & 85.0 & 70.0 & 80.0 & 70.0 & 65.0 & 10.0 & 10.0 & 10.0 & 0.0 \\
Claude Opus 4.5 & 90.0 & 80.0 & 80.0 & 80.0 & 65.0 & 70.0 & 10.0 & 20.0 & 10.0 & 0.0 \\
o4-mini & 65.0 & 85.0 & 90.0 & 60.0 & 55.0 & 70.0 & 75.0 & 5.0 & 0.0 & 0.0 \\
Gemini 3 Pro Preview & 80.0 & 65.0 & 85.0 & 65.0 & 70.0 & 65.0 & 10.0 & 5.0 & 0.0 & 0.0 \\
\midrule
\textbf{Average} & \textbf{85.0} & \textbf{83.3} & \textbf{79.2} & \textbf{76.7} & \textbf{70.8} & \textbf{66.7} & \textbf{41.7} & \textbf{26.7} & \textbf{8.3} & \textbf{0.8} \\
\midrule
\multicolumn{11}{l}{\textbf{Joint optimality (\%)}} \\
\midrule
GPT-5.2-pro & 0.0 & 90.0 & 25.0 & 70.0 & 55.0 & 25.0 & 0.0 & 40.0 & 0.0 & 0.0 \\
GPT-5.2 & 0.0 & 75.0 & 15.0 & 40.0 & 45.0 & 10.0 & 0.0 & 25.0 & 0.0 & 0.0 \\
Claude Opus 4.6 & 0.0 & 85.0 & 10.0 & 40.0 & 40.0 & 25.0 & 0.0 & 10.0 & 5.0 & 0.0 \\
Claude Opus 4.5 & 0.0 & 25.0 & 0.0 & 35.0 & 45.0 & 0.0 & 0.0 & 10.0 & 10.0 & 0.0 \\
o4-mini & 0.0 & 30.0 & 10.0 & 15.0 & 30.0 & 0.0 & 0.0 & 0.0 & 0.0 & 0.0 \\
Gemini 3 Pro Preview & 0.0 & 35.0 & 5.0 & 5.0 & 40.0 & 0.0 & 0.0 & 0.0 & 0.0 & 0.0 \\
\midrule
\textbf{Average} & \textbf{0.0} & \textbf{56.7} & \textbf{10.8} & \textbf{34.2} & \textbf{42.5} & \textbf{10.0} & \textbf{0.0} & \textbf{14.2} & \textbf{2.5} & \textbf{0.0} \\
\bottomrule
\end{tabular}
}
\end{table*}

\subsection{Feasibility-Optimality Decoupling}
\label{sec:exp:decoupling}

The clearest finding is the decoupling between feasibility and optimality in facility location and vehicle routing. Facility location has the highest average feasibility of any domain (85.0\%, with one model at 100\%) yet 0\% joint optimality across all six models. Vehicle routing shows a similar pattern for several models. GPT-5.2 achieves 80.0\% feasibility but 0\% optimality.

\Cref{tab:gap_distribution} presents the optimality gap distribution for feasible solutions across domains.

\begin{table}[ht]
\centering
\small
\caption{Optimality gap distribution for feasible solutions.}
\vspace{6pt}
\label{tab:gap_distribution}
\begin{tabular}{lrrrrr}
\toprule
Domain & Feasible Sol.\ & Within 0.1\% & Within 1\% & Within 5\% & Median Gap \\
\midrule
Production Mix & 100 & 68 (68\%) & 87 (87\%) & 96 (96\%) & 0.03\% \\
Shift Scheduling & 85 & 51 (60\%) & 62 (73\%) & 80 (94\%) & 0.00\% \\
Bin Packing & 32 & 17 (53\%) & 17 (53\%) & 20 (62\%) & 0.00\% \\
Job Shop & 92 & 41 (45\%) & 41 (45\%) & 64 (70\%) & 1.32\% \\
Project Planning & 10 & 3 (30\%) & 3 (30\%) & 8 (80\%) & 4.35\% \\
Order Fulfillment & 95 & 13 (14\%) & 44 (46\%) & 65 (68\%) & 1.16\% \\
Portfolio & 80 & 12 (15\%) & 34 (42\%) & 76 (95\%) & 1.11\% \\
Vehicle Routing & 50 & 0 (0\%) & 3 (6\%) & 20 (40\%) & 9.85\% \\
Facility Location & 102 & 0 (0\%) & 2 (2\%) & 28 (27\%) & 9.41\% \\
Crew Assignment & 1 & 0 (0\%) & 0 (0\%) & 0 (0\%) & 7.04\% \\
\bottomrule
\end{tabular}
\end{table}

Facility location illustrates this well. With 102 feasible solutions across all models and a median optimality gap of 9.41\%, models consistently find solutions that satisfy all constraints but are substantially suboptimal. The minimum gap observed across all models is 0.80\%, meaning no model ever came within 0.1\% of the Gurobi optimum despite the domain's relatively simple constraint structure (capacity, distance, and logical constraints). This suggests that models use reasonable heuristics (open cheap facilities, assign nearby customers) that produce good-but-not-optimal solutions. The exact optimum requires precise trade-offs between fixed costs, variable costs, and capacity utilization that heuristic reasoning does not capture.

Vehicle routing shows an even larger gap, with median 9.85\% among 50 feasible solutions, with only 3 solutions (6\%) within 1\% of optimal. Models can construct valid routes but struggle to optimize total distance, likely because finding short routes through multiple customers with time windows requires the kind of systematic search that branch-and-bound provides but autoregressive generation does not.

By contrast, production mix (median gap 0.03\%) and shift scheduling (median gap 0.00\%) show that for some problem structures, LLMs can find near-optimal solutions reliably once they achieve feasibility.

The structure of this gap, specifically which domains are straightforward versus difficult for optimization conditional on feasibility, provides a complementary difficulty axis to feasibility itself. The two axes are weakly correlated. Domains that are difficult for feasibility (crew assignment, project planning) tend to have too few feasible solutions to assess optimization quality, while some domains with high feasibility (facility location) exhibit large optimality gaps.

\subsection{Failure Mode Breakdown}
\label{sec:exp:failures}

We analyzed all 4{,}314 constraint violations across the 1{,}200 model-task evaluations (6 models $\times$ 200 tasks) and identified four systematic failure categories.

\textbf{Category 1, structural misunderstanding (project planning, crew assignment; 2{,}758 violations).} Project planning accounts for 1{,}412 violations dominated by ``duration too short'' errors across all models and nearly all tasks. Models systematically underestimate task durations, producing schedules where tasks are allocated fewer periods than required. This is not a failure on individual constraint reasoning but a systematic misparse of how duration constraints work in the domain. Crew assignment accounts for 1{,}346 violations across 38 unique violation types, including under-coverage errors (tasks not allocated enough hours) and workload balance violations. The diversity of violation types suggests models cannot reliably parse the problem structure.

\textbf{Category 2, entity hallucination (bin packing; 691 violations).} The dominant violations in bin packing are references to bins that do not exist in the problem (e.g., ``Unknown bin: b2-1''). Models hallucinate entity identifiers rather than using only the IDs provided in the prompt. This is a structural output error, a failure to ground the solution in the provided entities, rather than a reasoning error about constraint satisfaction.

\textbf{Category 3, constraint-specific reasoning failures (vehicle routing, portfolio, shift scheduling; 615 violations).} These are the most nuanced failures. Models demonstrate partial competence (they understand the domain structure) but fail on specific constraint types. In vehicle routing, the dominant violation (29 instances) is ``All routes must respect each customer's specified delivery time window.'' Models construct valid routes but ignore time windows. In portfolio optimization, violations cluster around group allocation constraints (e.g., ``Maintain at least 25\% in defensive correlation-group assets''). Models optimize the expected return but miss secondary allocation requirements. In shift scheduling, violations involve fatigue rules and rest requirements between consecutive shifts.

\textbf{Category 4, budget and capacity threshold violations (facility location, order fulfillment; 105 violations).} Facility location violations (79) are dominated by budget exceedances. Models open slightly too many facilities, pushing total fixed costs above the budget constraint. Order fulfillment violations (26) involve minimum order quantities and conditional vendor constraints. These are ``near-miss'' errors where solutions are structurally sound but violate resource bounds, often by small margins. The remaining 145 violations span miscellaneous constraint types that do not cluster into a single category.

\begin{figure}[ht]
\centering
\includegraphics[width=0.8\columnwidth]{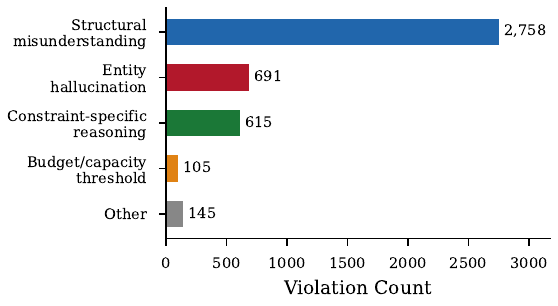}
\caption{Constraint violation breakdown by failure category.}
\label{fig:violations}
\end{figure}

The failure modes form a rough hierarchy of severity. Structural misunderstanding (Categories 1 and 2) reflects a fundamental failure to parse the problem correctly, while constraint-specific reasoning failures (Category 3) indicate partial competence with targeted weaknesses. Budget and threshold violations (Category 4) are the mildest; the solutions are largely correct but fail a quantitative check. This breakdown suggests that improving LLM constraint reasoning requires both better grounding in problem structure (for Categories 1 and 2) and more systematic constraint tracking during generation (for Categories 3 and 4).

\subsection{Self-Report Discrepancy}
\label{sec:exp:selfreport}

ConstraintBench's independent objective verification, where the verifier recalculates the true objective from raw decision variables rather than trusting the model's self-reported value, enables measurement of self-report discrepancy, the gap between what a model claims its objective is and what it actually achieves.

\begin{table}[ht]
\centering
\caption{Self-report discrepancy statistics.}
\vspace{6pt}
\label{tab:self_report}
\begin{tabular}{lrrr}
\toprule
Model & Discrepant / 200 & Median Error (\%) & Max Error (\%) \\
\midrule
GPT-5.2-pro & 22 & 13.7 & 128.9 \\
GPT-5.2 & 24 & 10.6 & 125.9 \\
Claude Opus 4.6 & 20 & 12.8 & 104.5 \\
Claude Opus 4.5 & 24 & 9.4 & 107.0 \\
o4-mini & 26 & 3.8 & 129.3 \\
Gemini 3 Pro Preview & 34 & 20.9 & 9{,}999{,}900.0 \\
\bottomrule
\end{tabular}
\end{table}

Most models report accurately on the majority of tasks, but all exhibit outliers where the reported objective is dramatically wrong. GPT-5.2-pro, for instance, produced a correct assignment on one task but reported an objective value 2.3$\times$ the true value (128.9\% error). Gemini 3 Pro Preview has several portfolio outliers where the model reported return rates as percentages rather than decimals (e.g., reporting 105.7 instead of 0.1057), with scaling errors ranging from a factor of 1{,}000 to over 100{,}000 in the most extreme case. These are unit-conversion errors rather than reasoning failures, but they illustrate that self-reported objectives can be wrong in systematic and domain-specific ways.

This finding has practical implications. In deployment settings where models assess their own solution quality (e.g., agentic loops that decide whether to accept or refine a solution), self-reported objectives cannot be trusted. Independent verification is necessary.

\section{Discussion}
\label{sec:discussion}

\subsection{Limitations}
\label{sec:discuss:limitations}

The benchmark comprises 200 tasks (20 per domain), which is small relative to some coding benchmarks such as SWE-bench~\cite{jimenez2024swebench} with 2{,}294 tasks. This reflects the cost of evaluating frontier models with extended reasoning budgets, where each evaluation costs \$0.50 to \$3.00 per task depending on the model. The per-domain sample of 20 is sufficient to identify the large effects we report (100$\times$ feasibility variation across domains) but insufficient for fine-grained analysis within domains. Each task is evaluated once, and while temperature-0 decoding ensures determinism (verified empirically by re-running 20 randomly selected tasks with identical outputs), this precludes variance estimates from sampling.

Ten domains provide broad but not exhaustive coverage of Operations Research. Notable omissions include network design, cutting stock, knapsack, and scheduling variants. We are extending to 100 domains in ongoing work, though only the original 10 have the full perturbation infrastructure.

The direct-solution framing is arguably not the most practical deployment mode for OR\@. In practice, formulation plus solver execution is more reliable for most applications. The direct-solution setting is scientifically interesting because it tests whether models have internalized constraint reasoning patterns, and it is practically relevant to settings where solver access is unavailable, where problems must be solved at low latency, or where models serve as warm-start heuristics for downstream optimization. But it should not be interpreted as the recommended deployment mode for production OR\@.

The 0.1\% optimality threshold is strict by design, and the absolute optimality numbers are sensitive to threshold choice. Facility location, for instance, has 85.0\% average feasibility but 0\% joint optimality at 0.1\%; at a 5\% threshold, it would rise to approximately 23\%. Production mix, by contrast, is threshold-insensitive (68\% of feasible solutions are within 0.1\%). We report threshold sensitivity analysis in \Cref{app:threshold} to enable readers to assess robustness of the headline numbers.

Finally, the ``errors'' column in \Cref{tab:aggregate} reflects cases where models failed to produce a parseable solution at all. Of the 22 total errors across all models, the large majority (18) are token exhaustion failures where the model spent its entire output budget on chain-of-thought reasoning and never produced a structured answer (stop\_reason=max\_tokens). The remaining errors are API connection failures and parse retries that did not converge. Claude Opus 4.6 accounts for 13 of the 22 errors, likely because its extended thinking consumed the token budget before producing output. Under our errors-as-infeasible methodology, these reduce its apparent feasibility. We report error counts alongside all metrics to enable alternative analyses.

\subsection{Future Work}
\label{sec:discuss:future}

We are extending ConstraintBench from 10 to 100 OR domains. Ninety additional domains, covering network design, cutting stock, workforce planning, supply chain optimization, and others, follow the same architecture of Pydantic schema, Gurobi model builder, independent verifier, and seed-based generation.

Beyond expanding domain coverage, we have developed a perturbation engine that leverages Gurobi's sensitivity analysis to generate problem variants from base instances. The engine operates in two modes. SAT-preserving perturbation modifies constraint parameters within solver-certified ranges to produce variant problems that remain feasible. UNSAT-generating perturbation uses binary search to find the feasibility boundary and produce controlled infeasible instances near the SAT/UNSAT tipping point. Combined with the seed-based generator, this infrastructure can produce hundreds of thousands of unique, solver-verified problem instances, enabling its use as a scalable training environment for reinforcement learning with deterministic, formally grounded reward signals~\cite{haladir2026rlfr}.

We are also developing a post-generation tightening mechanism that replaces loose constraint bounds with bounds derived from the optimal solution, creating calibrated difficulty levels. For example, a budget constraint can be set at $1.15 \times f(x^*)$ for cost-minimization domains, or a minimum-profit constraint at $0.93 \times f(x^*)$ for maximization domains. This would transform tasks where any feasible solution is near-optimal into tasks where only solutions close to the optimum satisfy all constraints, enabling finer-grained measurement of optimization quality. The current benchmark evaluates problems as generated, without such tightening.

Two additional directions are in progress. First, we plan to investigate scenario framing bias. Two problem instances may share identical constraint sets and objectives but differ in scenario context, such as an automotive parts distribution network versus a hospital supply chain. If models produce systematically different solutions for mathematically isomorphic problems wrapped in different narratives, this would reveal that LLM reasoning is influenced by surface-level framing in ways that exact solvers are not. Second, we are developing a multi-turn agentic evaluation mode where models receive direct access to a Gurobi solver instance and can iteratively formulate, solve, and refine their approach, testing whether models can use formal verification tools effectively rather than relying on single-pass constraint reasoning.

\section{Conclusion}
\label{sec:conclusion}

We introduced ConstraintBench, a benchmark for evaluating LLMs on direct constrained optimization across 10 Operations Research domains with all ground-truth solutions verified by the Gurobi Optimizer. Evaluating six frontier models on 200 tasks, we find that constraint satisfaction, not objective optimization, is the primary bottleneck. The best model achieves 65.0\% feasibility, and feasible solutions average 89 to 96\% of the Gurobi optimum, yet no model exceeds 30.5\% on joint feasibility and optimality. Per-domain analysis reveals a difficulty spectrum spanning two orders of magnitude, from near-solved (facility location, 85.0\% average feasibility) to near-impossible (crew assignment, 0.8\%), with systematic failure modes that range from structural misunderstanding of domain semantics to targeted constraint reasoning errors. The feasibility-optimality decoupling in facility location (85.0\% feasibility, 0\% optimality, median gap 9.41\%) and vehicle routing demonstrate that achieving feasibility and achieving optimality are distinct challenges that current models address unevenly. By providing solver-verified ground truth, constraint-level evaluation, and detailed failure diagnostics, ConstraintBench offers a rigorous measurement of LLM constraint reasoning and a clear target for improvement.


\newpage
\appendix
\section*{Appendices}

\section{Example Prompt}
\label{app:example_prompt}

The following is a representative (abbreviated) prompt from the facility location domain.

\begin{center}
\setlength{\fboxsep}{8pt}
\setlength{\fboxrule}{0.4pt}
\fbox{%
\begin{minipage}{0.95\linewidth}
\footnotesize\ttfamily
You are solving a facility location optimization problem.

SCENARIO: Automotive dealership network: RoadForge Auto Group must decide which regional Parts Distribution \& Reconditioning Centers to open to supply OEM parts, accessories, and certified pre-owned reconditioning kits to a network of dealerships. Each dealership must be served by exactly one open center within 50 distance units.

=== CUSTOMERS ===\\
RoadForge Downtown Motors (id: c1): at (12.0, 18.0), demand=70\\
RoadForge Riverfront Auto (id: c2): at (18.0, 12.0), demand=60\\
\ldots

=== POTENTIAL FACILITIES ===\\
GearPoint Central PDC (id: f1): at (16.0, 16.0), capacity=500, fixed\_cost=\$42000, var\_cost=\$3.6/unit\\
\ldots

=== ADDITIONAL CONSTRAINTS ===\\
GearPoint Central PDC must remain operational (existing OEM contract)\\
TorqueTown West and WrenchWorks Northwest cannot both operate (union jurisdiction)\\
If ChromeCove Airport is opened, PistonPort East must also be opened (linehaul dependency)\\
\ldots

=== YOUR TASK ===\\
Minimize total cost (fixed costs + variable costs) while serving every dealership within max\_distance, respecting capacity limits, and satisfying all logical constraints.
\end{minipage}
}
\end{center}

\section{Evaluation Hyperparameters}
\label{app:eval_hyperparams}
\vspace{-4pt}
\begin{table}[ht]
\centering
\footnotesize
\caption{Evaluation hyperparameters.}
\label{tab:hyperparams}
\setlength{\tabcolsep}{4pt}
\begin{tabular}{ll}
\toprule
Parameter & Value \\
\midrule
Temperature & 0 \\
Max output tokens & 64{,}000 \\
Reasoning tokens (OpenAI) & 55{,}000 (effort: high) \\
Extended thinking (Anthropic) & Enabled, same budget \\
Structured output & Enabled (all providers) \\
Trials per task & 1 \\
Optimality threshold ($\epsilon$) & 0.1\% \\
\bottomrule
\end{tabular}
\end{table}

\section{Provider-Specific Evaluation Configuration}
\label{app:hyperparams}
\vspace{-4pt}

\begin{table}[ht]
\centering
\footnotesize
\caption{Provider-specific evaluation configuration.}
\label{tab:provider_config}
\setlength{\tabcolsep}{3pt}
\begin{tabular}{llll}
\toprule
Provider & Structured Output Mode & Reasoning Config & Retry on Parse Failure \\
\midrule
OpenAI (GPT-5.2, GPT-5.2-pro) & Responses API, structured output & reasoning\_effort=high, 55K tokens & 3 retries \\
OpenAI (o4-mini) & Responses API, structured output & reasoning\_effort=high, 55K tokens & 3 retries \\
Anthropic (Claude Opus 4.5, 4.6) & Structured schema & Extended thinking enabled, same budget & 3 retries \\
Google (Gemini 3 Pro Preview) & JSON mode & Default reasoning & 3 retries \\
\bottomrule
\end{tabular}
\end{table}

\newpage
\section{Threshold Sensitivity}
\label{app:threshold}
\vspace{-4pt}

\begin{table}[ht]
\centering
\footnotesize
\caption{Average \% optimal by threshold.}
\label{tab:threshold}
\setlength{\tabcolsep}{4pt}
\begin{tabular}{lrrrr}
\toprule
Domain & 0.1\% & 0.5\% & 1.0\% & 5.0\% \\
\midrule
Production Mix & 56.7 & 64.2 & 72.5 & 80.0 \\
Facility Location & 0.0 & 0.0 & 1.7 & 23.3 \\
Job-Shop Scheduling & 34.2 & 34.2 & 34.2 & 53.3 \\
Order Fulfillment & 10.8 & 15.0 & 36.7 & 54.2 \\
Shift Scheduling & 42.5 & 44.2 & 51.7 & 66.7 \\
Portfolio & 10.0 & 21.7 & 28.3 & 63.3 \\
Bin Packing & 14.2 & 14.2 & 14.2 & 16.7 \\
Vehicle Routing & 0.0 & 0.8 & 2.5 & 16.7 \\
Project Planning & 2.5 & 2.5 & 2.5 & 6.7 \\
Crew Assignment & 0.0 & 0.0 & 0.0 & 0.0 \\
\midrule
\textbf{Aggregate} & \textbf{17.1} & \textbf{19.7} & \textbf{24.4} & \textbf{38.1} \\
\bottomrule
\end{tabular}
\end{table}

\end{document}